\documentclass[letterpaper, 10 pt, conference]{ieeeconf}
\usepackage{amsmath,amssymb,amsfonts}
\usepackage{siunitx}
\usepackage{todonotes}
\usepackage{cite}
\usepackage{graphicx}
\usepackage{algorithm}
\usepackage{hyperref}
\usepackage{xcolor}
\usepackage{titlesec}
\usepackage{algpseudocode}
\usepackage[belowskip=-4pt,aboveskip=2pt, font=small]{caption}
\usepackage{subcaption}
\definecolor{green}{RGB}{225,128,20}

\IEEEoverridecommandlockouts                              

\title{\LARGE \bf
Focused Adaptation of Dynamics Models for Deformable Object Manipulation
}

\author{Peter Mitrano$^{1}$, Alex LaGrassa$^{2}$, Oliver Kroemer$^{2}$, and Dmitry Berenson$^{1}$
\thanks{$^{1}$Department of Robotics, University of Michigan}%
\thanks{$^{2}$Robotics Institute, Carnegie Mellon University}%
\thanks{This work was supported in part by the Office of Naval Research Grants N00014-21-1-2118 and N00014-18-1-2775, NSF grants IIS-1956163, CMMI-1925130, IIS-1750489 and IIS-2113401, and ARL grant W911NF-18-2-0218.}%
}

\begin{document}

\maketitle
\thispagestyle{empty}
\pagestyle{empty}

\newcommand{\FOCUS}{FOCUS}

\begin{abstract}

In order to efficiently learn a dynamics model for a task in a new environment, one can adapt a model learned in a similar source environment. However, existing adaptation methods can fail when the target dataset contains transitions where the dynamics are very different from the source environment. For example, the source environment dynamics could be of a rope manipulated in free space, whereas the target dynamics could involve collisions and deformation on obstacles. Our key insight is to improve data efficiency by focusing model adaptation on only the regions where the source and target dynamics are similar. In the rope example, adapting the free-space dynamics requires significantly less data than adapting the free-space dynamics while also learning collision dynamics. We propose a new method for adaptation that is effective in adapting to regions of similar dynamics. Additionally, we combine this adaptation method with prior work on planning with unreliable dynamics to make a method for data-efficient online adaptation, called \FOCUS{}. We first demonstrate that the proposed adaptation method achieves statistically significantly lower prediction error in regions of similar dynamics on simulated rope manipulation and plant watering tasks. We then show on a bimanual rope manipulation task that \FOCUS{} achieves data-efficient online learning, in simulation and in the real world.

\end{abstract}

\section{INTRODUCTION}
Autonomous systems often rely on a dynamics model, which predicts future states given actions, to reach a desired goal state. However, general and accurate dynamics models only exist for a narrow range of robotics problems. Learning these dynamics models is an increasingly popular paradigm, in part because learned models can be repeatedly improved using autonomously collected real-world data. However, fine-tuning an initial dynamics model on new data can perform poorly when the data contains complex dynamics on which the dynamics model was not initially trained. For example, suppose we want to manipulate a rope amongst clutter, and we have a dynamics model trained on free-space motions in simulation. Free-space transitions in the real world are fairly similar to free-space transitions in simulation, but transitions where the rope deforms on objects in the scene are very different from anything seen in simulation. We call these transitions \emph{distracting}, because they are hard to learn from a few examples, and because they make it harder to adapt accurately to the free-space dynamics. More generally, transitions from regions of dissimilar dynamics can inhibit effective transfer to regions of similar dynamics. This problem is similar to ``cleaning'' data in machine learning~\cite{mislabeled99,filtering21,anomoly22}. For dynamics learning, defining what ``clean'' means can be difficult, and has not been studied extensively. Instead, the dominant paradigm is simply to train on all the collected data.

However, training on all the data can fail because real-world datasets for learning dynamics are often too small to learn generalized models over the entire state-action space. In our experiments, we show that simply fine-tuning on all the data can yield a model that is not accurate enough for planning. If the task can be completed while remaining in regions where dynamics are similar, then it can be worth trading accuracy in dissimilar regions for accuracy in similar regions.
Our key insight is that, when we are adapting from an initial model, we can leverage the initial model to achieve significantly lower prediction error by focusing on transitions where the source and target dynamics are the most similar. The idea that transfer is easier when the source and target data are similar is well-supported in the transfer learning literature~\cite{sorocky2020experience,bocsi2013alignment}.

\begin{figure}
    \centering
    \includegraphics[width=1\linewidth]{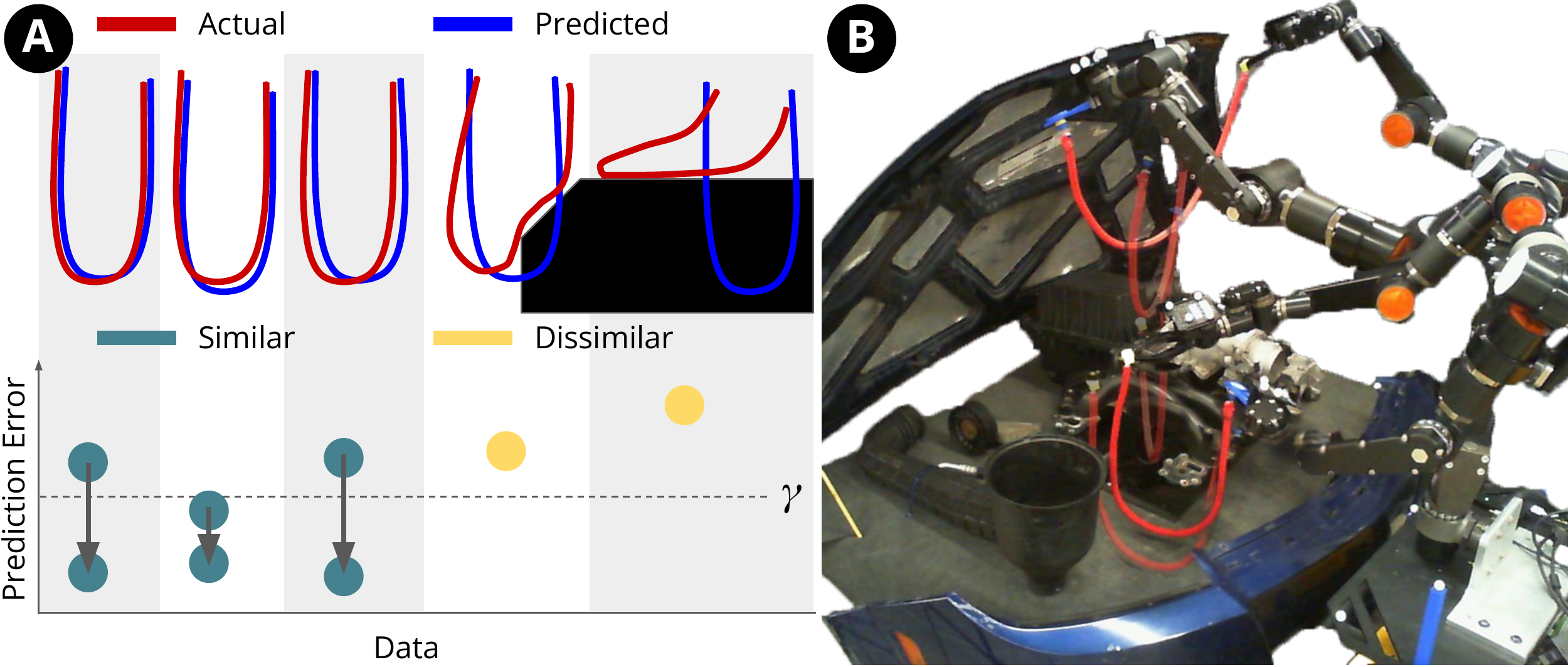}
    \caption{
    (A) An illustration of how our adaptation method focuses on regions where the source and target dynamics are similar. When focusing adaption on free-space dynamics, the prediction errors decrease for other free-space data (similar), and do not decrease for collision dynamics (dissimilar).
    (B) A mock-up of a car engine bay. The robot must move the rope and place it under the engine without snagging it to set up for lifting the engine. We use our proposed adaptation method to improve the success rate during online learning for this task.}
    \label{fig:real_robot_setup}
\end{figure}

To implement this strategy, we propose an adaptation method that minimizes prediction error in regions where the source and target dynamics are similar. The proposed method minimizes prediction error in these regions by fine-tuning on an initially small set of data from these regions, and growing that subset over the course of training. This is done with a loss function inspired by curriculum learning that weights transitions according to their prediction error, assigning a higher weight to low-error transitions. Under the assumption that there are paths to the goal where the source and target dynamics are similar, this adaptation method can be used to achieve high task success in the target environment.

The first contribution of this paper is a method for adapting dynamics models to datasets that contain distracting transitions. We demonstrate that the proposed method is successful in filtering out distracting data and that the resulting model is more accurate in the regions of state-action space where the source and target dynamics are similar. The second contribution is a data-efficient online-learning method that pairs our adaptation method with prior work on planning with unreliable dynamics models~\cite{UnreliableMitrano2021,MDEs22}. We call our combined method for online learning \FOCUS{}. \FOCUS{} achieves higher success rates in the low-data regime because the adapted dynamics are more accurate, which leads to finding more reliable plans.

\section{RELATED WORK}

\textbf{Adapting dynamics models: } One approach to adaptation is to use system identification to fit a global dynamics model of the target domain~\cite{sastry1989adaptive, arndt2021few, evans2022context, nagabandi19learning}. Alternatively, domain randomization uses random variations of conditions during training to make the model robust to that type of variance during test time~\cite{lowrey2018reinforcement}. Some methods go further, and iteratively refine the noise distribution by comparing simulations to rollouts executed in the real world~\cite{chebotar2019closing, langsfeld2018selection}. These methods require knowledge of how parameters can vary between the source and the target domains. By contrast, our approach needs no knowledge of how the system parameters such as dynamical parameters, object geometry, and kinematics may vary. Prior work has also used estimates of similarity between the source and target system to guide adaptation, for instance by selecting training data most similar to the source system~\cite{sorocky2020experience} or selecting the most useful source domain for a given target domain~\cite{courchesne2021onassessing}. Similarly, our approach focuses both training and data collection on transitions with low prediction error, which are easier to learn. Previous work avoids negative transfer of dynamics from the source domain that is harmful to performance in the target domain \cite{sorocky2020experience, torrey2010transfer}. In contrast, our work focuses on avoiding distracting dynamics in the target domain that inhibit effective transfer to dynamics where the source and target are similar.

\textbf{Data Cleaning: } Data Cleaning is a term for methods that aim to remove training examples that are harmful, incorrect, or unhelpful~\cite{dataCleaningSurvey16}. Methods such as Majority Vote Filtering or the Iterative-Partitioning Filter can remove mislabeled classification data, but they do not apply to regression problems and assume that the type of data we want to fit is the majority type \cite{IPF07,mislabeled99, filtering21}. In contrast, learning dynamics is a regression problem, and in our experiments, the data we want to adapt to may be a minority.

\textbf{Curriculum learning: } Transfer learning methods work best between similar source and target domains~\cite{sorocky2020experience, torrey2010transfer}. To tackle larger differences, curriculum learning methods use intermediate problems through mechanisms such as pseudo-labels or task selection, which has been successful in classification, reinforcement learning, and machine translation~\cite{gradualAdaptation20, zhang2017curriculum, zhan2021meta, curriculumDeepRL20}. Like curriculum learning, our proposed method tries to train on easier data first, and keep the difference between the old and new data small~\cite{curriculumBengio09}. However, unlike in standard curriculum learning, we do not necessarily converge to training on all data.

\textbf{Planning with unreliable dynamics models: } If a dynamics model has a known probabilistic transition model, belief-space planning can be used~\cite{kaelbling2013integrated, platt2010belief}. Since meaningful predictive uncertainty distributions are difficult to estimate for novel scenarios, some methods instead directly predict model error and use it as a constraint for planning~\cite{MDEs22,UnreliableMitrano2021,UnreliableDale2019}.
Rather than avoiding uncertainty, model-based reinforcement learning methods try to reduce uncertainty by collecting new data and fine-tuning the dynamics~\cite{PILCO, nagabandi19learning}. In this work, we both adapt the dynamics online and avoid inaccurate predictions, which we find results in higher task success rates with less data.

\textbf{Rope Manipulation and Plant Watering: } Prior model-based rope manipulation and plant watering methods optimize parameters of a simulator or neural network to match real-world dynamics, as we do here. \cite{langsfeld2018selection} also uses a mechanism to avoid including data that is high error for a particular local model. Typically, the goal is to learn a model that is accurate for a wide range of dynamics for a task, without prioritizing dynamics that may require less data to learn~\cite{yan2021learning, seita2021learning, li2018learning, guevara2017adaptable, noda2007modeling, tsuji2014high, li2022graph, OfflineOnline22}. Since linear deformable objects are so high-dimensional, more work explores methods to learn or adapt models for a subset of configurations~\cite{UnreliableDale2019,  UnreliableMitrano2021,GlobalAdapt22}.

\section{METHODS}

First, we formalize the dynamics adaptation problem studied in this paper. Then, we describe our method for adapting the dynamics. Finally, we describe how our adaptation method can be used in an online adaptation framework to efficiently learn a rope manipulation task.

\subsection{Problem Statement}

\newcommand{\state}{s}
\newcommand{\action}{a}
\newcommand{\env}{\mathcal{E}}
\newcommand{\dynamics}{f}
\newcommand{\MDE}{h}
\newcommand{\pred}[1]{\hat{#1}}
\newcommand{\dynamicsErrort}{||\pred{\state}^t-\state^t||^2}
\newcommand{\dist}{d}
\newcommand{\mdeError}{d}
\newcommand{\source}[1]{#1_S}
\newcommand{\target}[1]{#1_T}
\newcommand{\learned}[1]{\hat{#1}}
\newcommand{\sourceDynamics}{\source{\dynamics}}
\newcommand{\targetDynamics}{\target{\dynamics}}
\newcommand{\learnedDynamics}{\learned{\dynamics}}
\newcommand{\initialDynamics}{\learnedDynamics_0}
\newcommand{\learnedDynamicsDef}{\pred{\state}'=\learnedDynamics(\state,\action)}
\newcommand{\dataset}{\mathcal{D}}
\newcommand{\DST}{\dataset_{ST}}
\newcommand{\goal}{\mathcal{G}}
\newcommand{\traj}{\{\state^0,\action^0,\dots,\action^{T-1},\state^T\}}
\newcommand{\softFilteringThrehsold}{\gamma}
\newcommand{\globalStep}{j}
\newcommand{\dmax}{d_\text{max}}
\newcommand{\transition}{(\state,\action,\state')}
\newcommand{\kMDE}{k_\MDE}

The problem addressed in this paper is to adapt a dynamics model trained in a source environment to data collected in a target environment, where the source and target environments have dynamics that are similar in some regions of the state-action space, but different in others. Furthermore, we consider the case where data collection is done by planning and executing paths to goals in the target environment using the learned dynamics model.

To formalize this, first consider the standard dynamics learning problem with a dataset $\dataset$ of transitions of states, actions, and next states $\transition$. We also assume a distance function $\dist(\state_1,\state_2)$ that returns a scalar is given, and that the state is fully observable. The true dynamics are $\state'=\dynamics(\state,\action)$, and the learned dynamics are $\pred{\state}'=\learnedDynamics(\state,\action)$. The initial dynamics $\initialDynamics$ is the model that is pre-trained in the source environment and not yet adapted to the target environment.  We define the source and target environment dynamics as similar for a transition if $\dist\big(\sourceDynamics(\state,\action), \targetDynamics(\state,\action)\big) < \softFilteringThrehsold$. The threshold $\softFilteringThrehsold$ should be small enough that it excludes distracting transitions, but large enough to include as much data from the target environment as possible. Let $\DST$ be the set of transitions from the target environment where this similarity condition holds.

In order to minimize the amount of data needed for adaptation to generalize, we aim to adapt the dynamics only to transitions from regions where this similarity condition holds ($\DST$). However, we also care about successfully completing the task, and therefore we also assume there are paths $\traj$ to the goal $\state^T\in\goal$ within the regions of similar dynamics $(\state^t,\action^t,\state^{t+1}) \in \DST $. This also means starting states are assumed to be in $\DST$. If this is not the case, a different source dynamics model may be needed.

While the ultimate objective of adaptation should be to maximize task success in the target environment, an important condition for task success is minimizing prediction error on $\DST$. If the goal is reachable within $\DST$ (as we assume), the prediction error on $\DST$ is small (our objective), and our motion planner is constrained to stay in $\DST$, then we can also expect high task success. Next, we discuss how our adaptation method minimizes error on $\DST$, followed by how we can achieve high task success by additionally constraining a motion planner to $\DST$.

\begin{figure}
    \centering
    \includegraphics[width=1\linewidth]{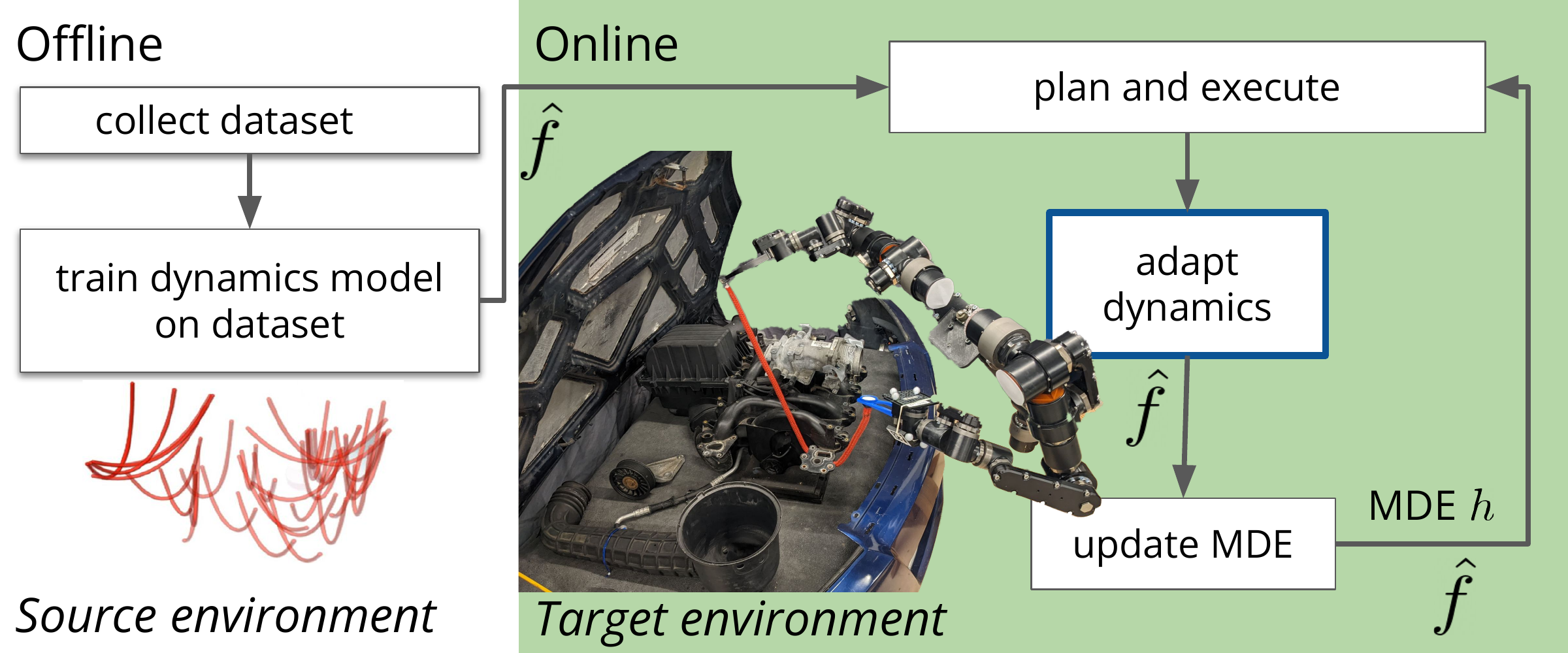}
    \caption{Block diagram showing the steps of our full online adaptation method. A dynamics model is initialized offline in the source environment (left), then adapted online in the target environment.}
    \label{fig:diagram}
\end{figure}

\subsection{Adapting the Dynamics}

Our objective is to minimize prediction error on $\DST$. At a high level, our method dynamically weights the training data $\dataset$ such that transitions that are likely to be in $\DST$ are given weights near 1, and transitions unlikely to be in $\DST$ are given weights near 0. Given a transition from our training data $\transition\in\dataset$, we cannot directly evaluate whether that transition is in $\DST$, since that would require knowing the true dynamics $\targetDynamics$. However, since the initial dynamics $\initialDynamics$ is trained to have low prediction error in the \textit{source} environment, transitions from the \textit{target} environment which also have low prediction error $\dist\big(\initialDynamics(\state,\action),\state'\big) < \softFilteringThrehsold$ are therefore in $\DST$. By training on transitions with initially low error, we expect the prediction error on other transitions which belong to $\DST$ to also decrease. This slowly brings more and more transitions to have prediction errors below $\softFilteringThrehsold$. On the other hand, the prediction error is unlikely to decrease for transitions not in $\DST$, because they are dissimilar to the transitions with low initial error.

Thus, at each step $j$ of training, we assign each transition a weight as a function of the prediction error $||\pred{\state}'-\state'||^2$ and multiply this weight by the loss. The full loss is shown in Equation \eqref{eq:adaptation_loss}.

\begin{equation}
    \label{eq:adaptation_loss}
    \begin{split}
    \mathcal{L}_\dynamics &= \frac{1}{T}\sum_{t=1}^T\Big(\dynamicsErrort w^t\Big) \\
    w^t &= 1 - \sigma\big(\phi(\globalStep)(\dynamicsErrort - \softFilteringThrehsold)\big)
    \end{split}
\end{equation}

When $\phi(\globalStep)$ is higher, the sigmoid $\sigma$ makes the boundary harder. For example, we use $\phi(j)=0.5j$ and $\phi(j) = 5j+3$ in our experiments. When $\globalStep$ is large, the boundary is almost hard, and transitions with error below $\softFilteringThrehsold$ have a weight near 1 and transitions with error above have a weight near 0. When $\globalStep$ is small, the boundary is soft, and the weights vary less. We found that allowing the weighting to be soft during early training steps improves the stability in the case where few or no transitions have errors below $\softFilteringThrehsold$ at the beginning of training. Adding a constant term ensures that $\phi(j) > 0$, which starts training with a harder boundary, reducing catastrophic forgetting at the beginning. The parameter $\softFilteringThrehsold$ can be chosen based on either the maximum error that can be corrected by a low-level controller, or based on the distribution of error on a validation set from the source environment (e.g. the 97th percentile).

\subsection{Online Learning}

\begin{figure}
    \centering
    \includegraphics[width=\linewidth]{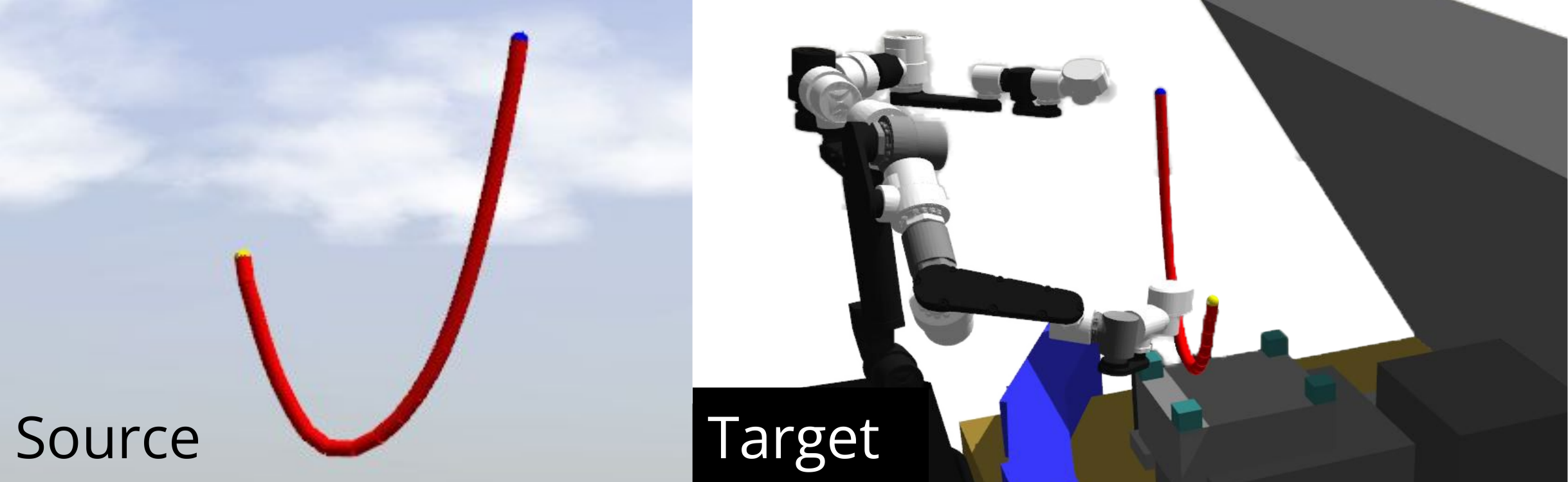}
    \caption{Source environment for bimanual rope manipulation (left) and simulated target environment (right) where there is a robot, obstacles, and the rope damping and stiffness are changed.}
    \label{fig:sim_rope_envs}
\end{figure}

\begin{figure}
    \centering
    \includegraphics[width=\linewidth]{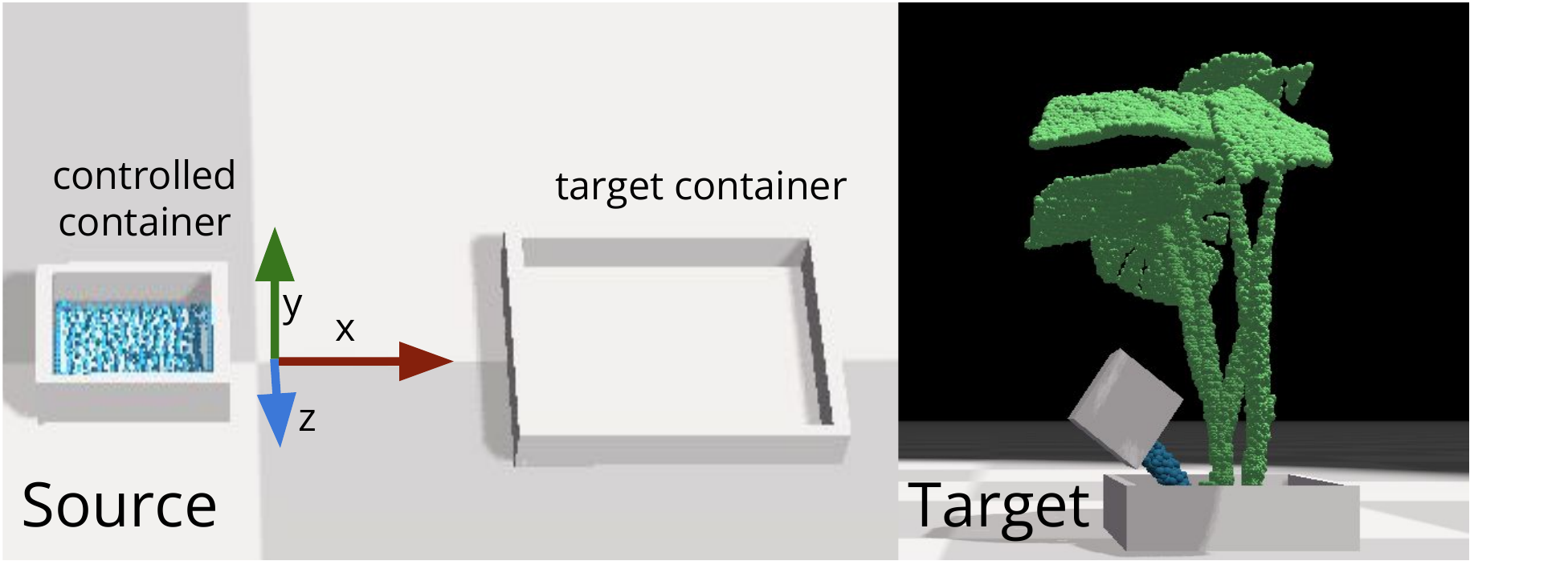}
    \caption{Source environment for plant watering (left) and target environment (right) where there is an additional plant, and the viscosity is tripled.}
    \label{fig:waterscene}
\end{figure}

In this section, we describe how the proposed adaptation method can be combined with prior work on planning with unreliable dynamics to achieve data-efficient online adaptation of dynamics models. A block diagram of the full method, which we call \FOCUS{}, is shown in Figure \ref{fig:diagram}. \FOCUS{} consists of an offline phase and an online phase. In the offline phase, we train a dynamics model using data from the source environment, which in our experiments is a simple simulation. We use random actions to collect a diverse set of data and standard techniques for training the fully-connected neural network dynamics model~\cite{UnreliableMitrano2021}.

In the online phase, we adapt the learned dynamics model to the target environment (e.g. the real world). This process alternates between (1) collecting new data in the target environment by planning and executing, (2) fine-tuning the dynamics, and (3) fine-tuning the model deviation estimator (MDE). We now explain the data collection and MDE fine-tuning steps.

\begin{figure}
    \centering
    \includegraphics[width=\linewidth]{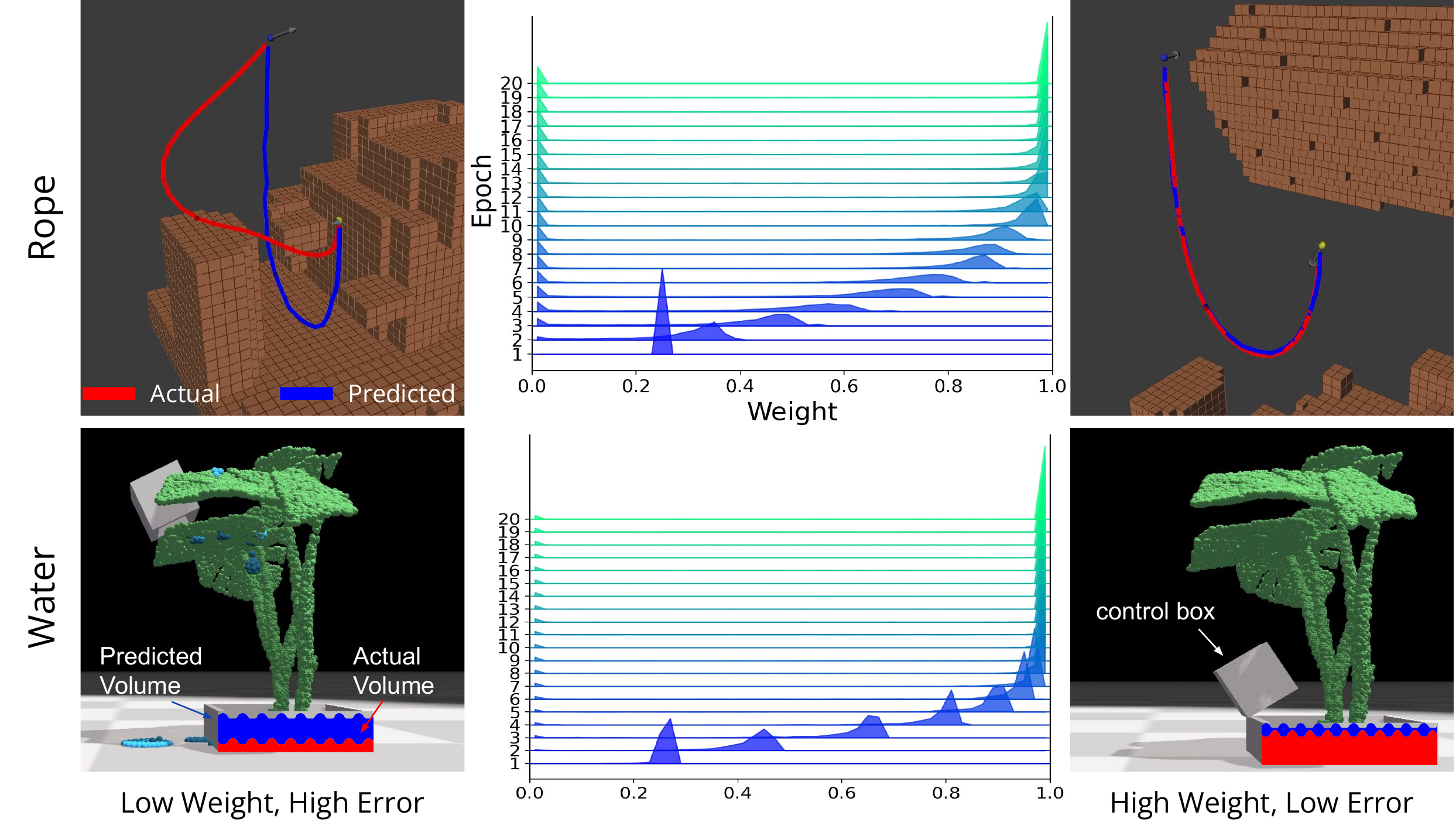}
    \caption{(center) Histograms showing weights assigned to the data according to Equation \eqref{eq:adaptation_loss} during the first 20 epochs of training. A histogram is shown for each epoch, where color varies with epoch, and these histograms are staggered along the y-axis. Initially, the weights vary only slightly across the data, but the distribution becomes strongly bimodal as training progresses. Examples of transitions given weight 0 (left) and weight 1 (right) at the end of training.}
    \label{fig:weights}
\end{figure}

\subsubsection{Planning and Execution for Data Collection}

We use a kinodynamic RRT planner where nodes are propagated using the learned dynamics model. Since the learned dynamics are adapting to only $\DST$ and are not accurate everywhere, we additionally constrain the planner to stay in $\DST$. Since $\DST$ is not known \textit{a priori}, we train another neural network, called a model deviation estimator (MDE)~\cite{MDEs22, UnreliableDale2019, UnreliableMitrano2021}, to predict the error of the dynamics model (more details in Section \ref{sec:fine_tuning_mde}). In addition to producing more robust plans, planning with MDEs has the additional benefit of focusing data collection on $\DST$. By collecting more data where the source and target dynamics are close, a larger fraction of $\dataset$ collected by the adaptation procedure is likely to be in $\DST$ and more useful for training.

We use the MDE in planning as a constraint. If the dynamics error predicted by the MDE is below a threshold $\dmax$, then we add it to the planning tree. We also randomly accept transitions with high predicted dynamics error with low probability (0.01), so that the planner will occasionally return paths with exploratory actions (we call these \textit{random-accepts}). These exploratory actions are essential for training the MDE since they can correct over-estimation of model error from the MDE. The threshold $\dmax$ for allowable error is similar to $\softFilteringThrehsold$ but may be set higher or lower to control the exploration/exploitation tradeoff.

The robot uses the planner to attempt the task, and repeatedly plans and executes open-loop until a timeout or the goal is reached. If no plan is found that reaches the goal, the plan which gets closest to the goal is executed. This repeated planning and execution is called one episode. After some fixed number of episodes (e.g. 10) we fine-tune the dynamics and the MDE using all data collected so far during the online phase.
\subsubsection{Fine-tune MDE}
\label{sec:fine_tuning_mde}

The MDE is used to constrain planning to regions where the dynamics model is predicted to be accurate, which has three benefits. First, we do not know the true error during planning, so the main purpose of the MDE is to give an estimate of error. Second, it helps bias data collection to contain transitions from $\DST$. Third, it makes reaching the goal more likely since it avoids plans that do not match the true dynamics. The MDE $\pred{\mdeError}=\MDE(\env,\state,\action,\pred{\state}')$ is a convolutional neural network that takes as input the environment, state, action, and next predicted state, and predicts the error of the dynamics model $\pred{\mdeError}$. We represent the environment $\env$ as a voxel grid of the static scene. The ground truth error used for training is the error between the true observed state and the state predicted by the learned dynamics: $d=\dist(\pred{\state}', \state')$. The loss function is shown in Equation \eqref{eq:mde_loss}. Intuitively, the MDE should be easier to learn with less data than learning the dynamics accurately everywhere, since the MDE need only predict the magnitude of the error as opposed to the full state vector~\cite{UnreliableMitrano2021}.
\begin{equation}
    \label{eq:mde_loss}
    \mathcal{L}_{\MDE} = ||\pred{\mdeError} - \mdeError||^2e^{-{\kMDE \mdeError}}
\end{equation}

$\kMDE$ is a hyperparameter that reduces the need to predict high dynamics errors with high accuracy. We set $\kMDE=10$.

\section{RESULTS}

We begin by describing the two domains for our experiments: bimanual rope manipulation and plant watering. We then validate the claims that (1) our proposed adaptation method achieves lower prediction error in regions of similar dynamics, and (2) that \FOCUS{} achieves higher success rates more quickly in the online adaptation setting compared to baselines that train on all data equally.

\begin{figure}
    \centering
    \includegraphics[width=\linewidth]{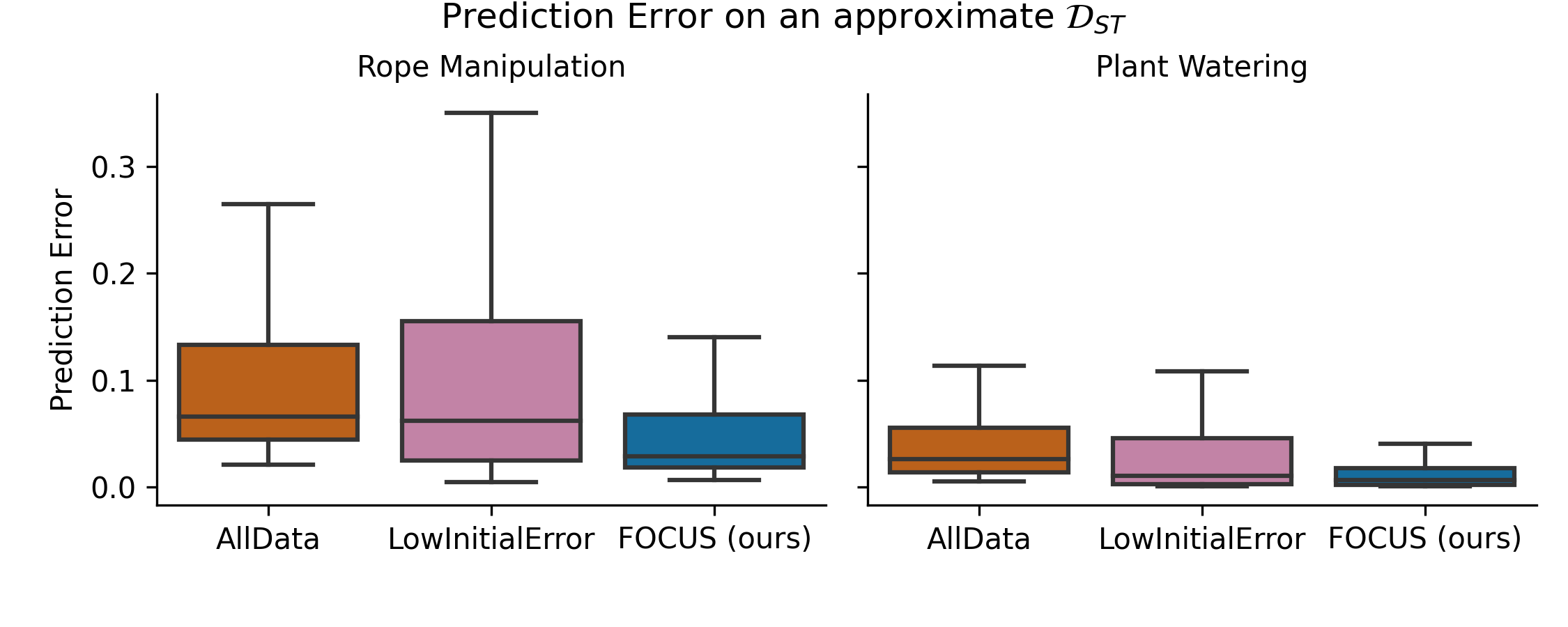}
    \caption{Prediction error for our method versus two baselines, evaluated on a dataset of transitions from regions where the source and target dynamics are similar.}
    \label{fig:validating}
\end{figure}

\textbf{Bimanual Rope Manipulation: }
In this task, a 16-DOF dual-arm robot is holding two ends of a rope in a scene resembling the engine bay of a car (scene shown in Figures \ref{fig:real_robot_setup},\ref{fig:sim_rope_envs}). The task mimics putting on lifting straps on the engine, which requires moving the rope through narrow passages and around protrusions. The goal is to place the middle of the rope in a goal region defined as a sphere of radius \SI{0.045}{\meter}. The planner outputs gripper position actions, and a local controller executes the actions while maintaining gripper orientations. The learned dynamics model predicts the state of the rope, represented as 25 points in 3D, given the initial rope state and gripper position actions. The distance function is the L2 norm of all rope points.

In the rope manipulation experiments, the source simulation has no obstacles and the robot is simplified to floating kinematic grippers. We then test adaptation to two different target environments: (1) another simulation that includes the robot and obstacles and has different damping and stiffness parameters for the rope, and (2) the real world. This tests adaptation to a different rope despite the distracting transitions where the rope deforms on the robot or the obstacles. Gazebo with ODE physics is used for simulation~\cite{gazebo}. For rope manipulation, the set $\DST$ would be the transitions from the target environment where the rope is in free space. We use $\softFilteringThrehsold=0.08$.

\begin{figure*}
     \centering
     Rope Manipulation \\
     \begin{subfigure}[b]{0.32\textwidth}
         \centering
         \includegraphics[width=\linewidth]{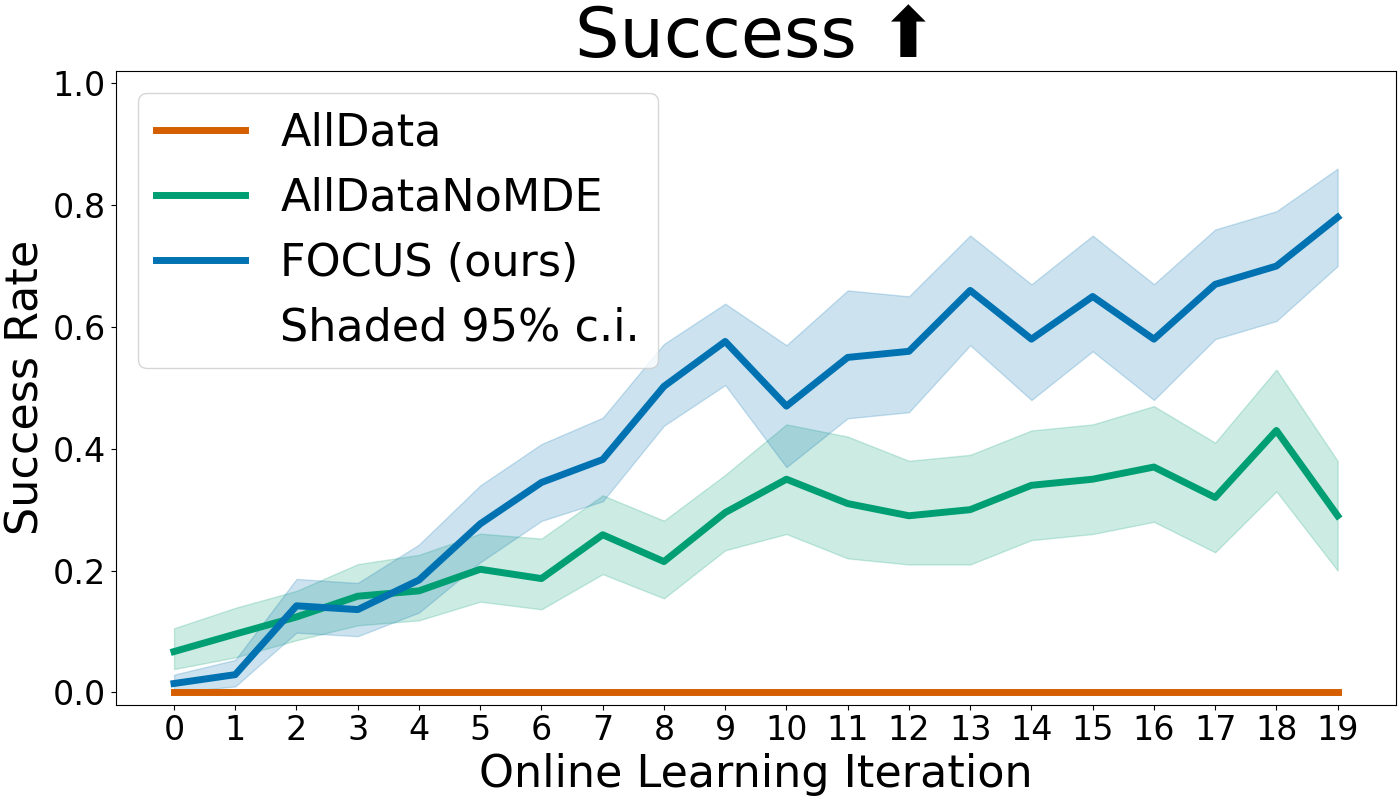}
     \end{subfigure}
     \hfill
     \begin{subfigure}[b]{0.32\textwidth}
         \centering
         \includegraphics[width=\linewidth]{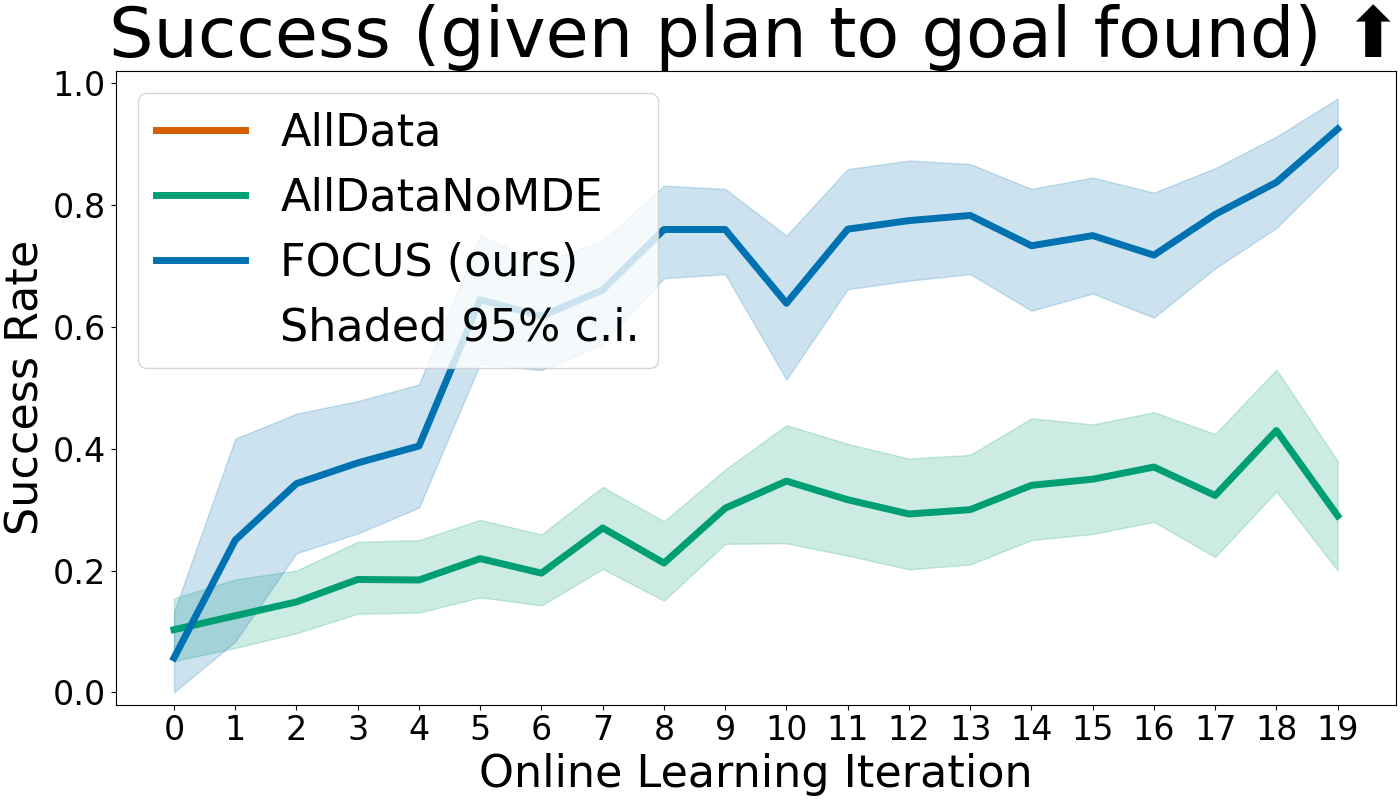} 
     \end{subfigure}
     \hfill
     \begin{subfigure}[b]{0.32\textwidth}
         \centering
         \includegraphics[width=\linewidth]{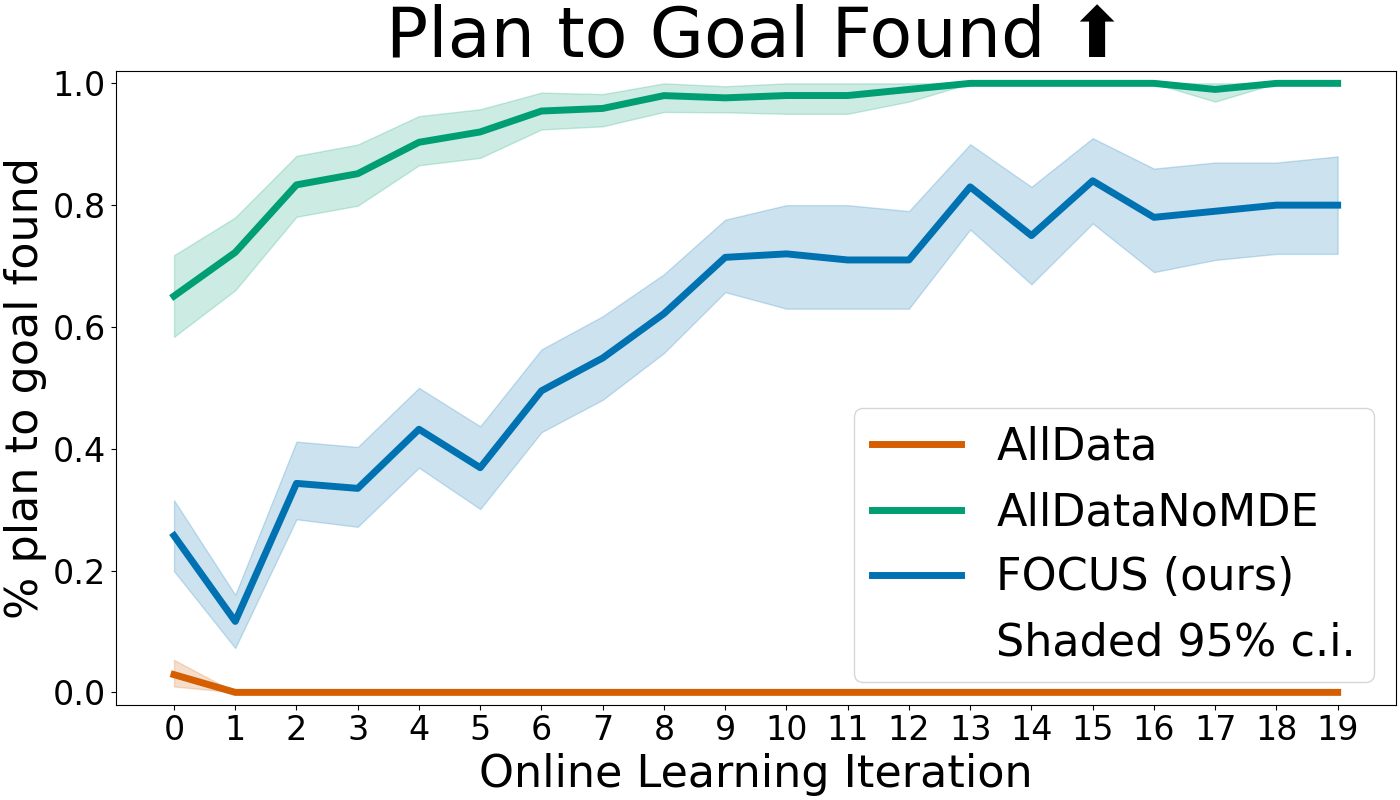}
     \end{subfigure}
     
    \noindent Plant Watering  \\
     \begin{subfigure}[b]{0.32\textwidth}
         \centering
         \includegraphics[width=\linewidth]{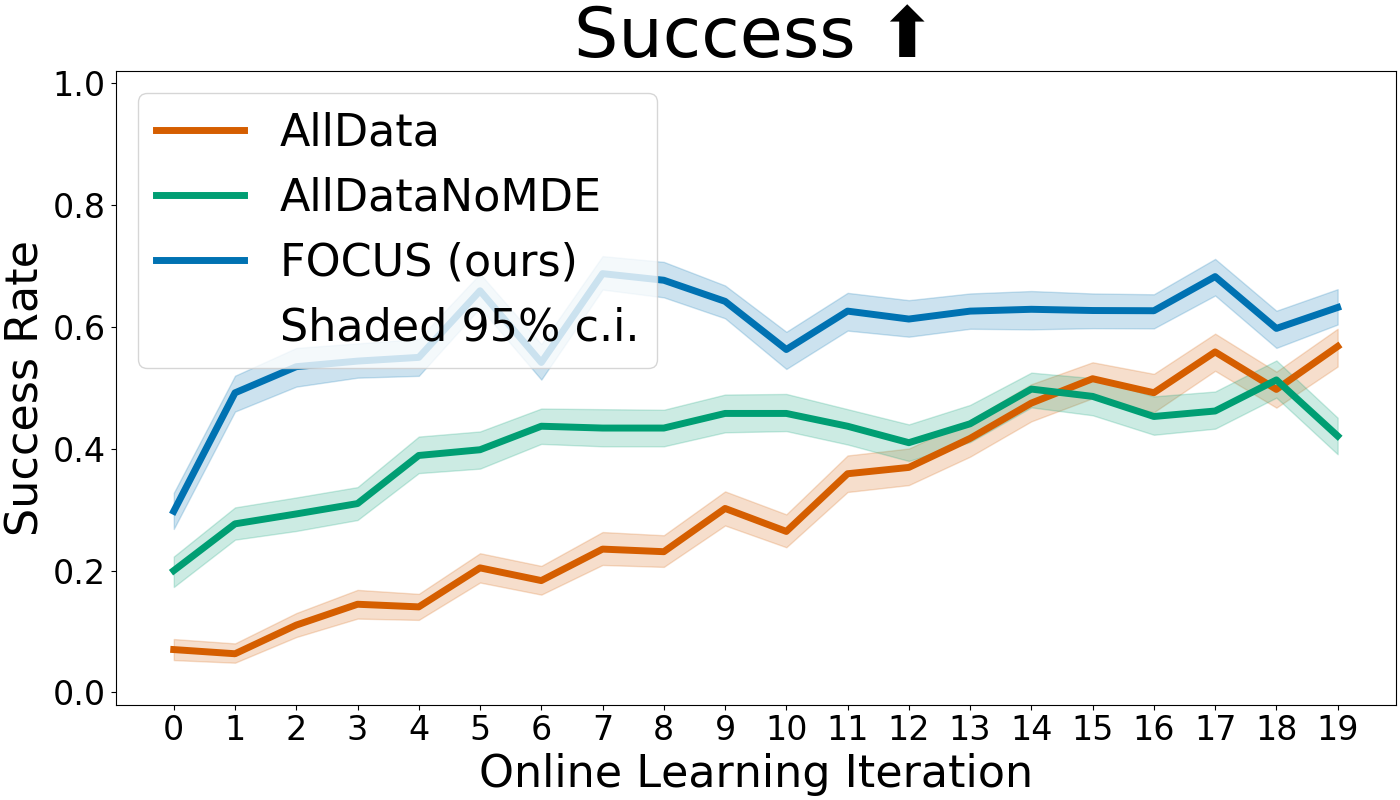}
     \end{subfigure}
     \hfill
     \begin{subfigure}[b]{0.32\textwidth}
         \centering
         \includegraphics[width=\linewidth]{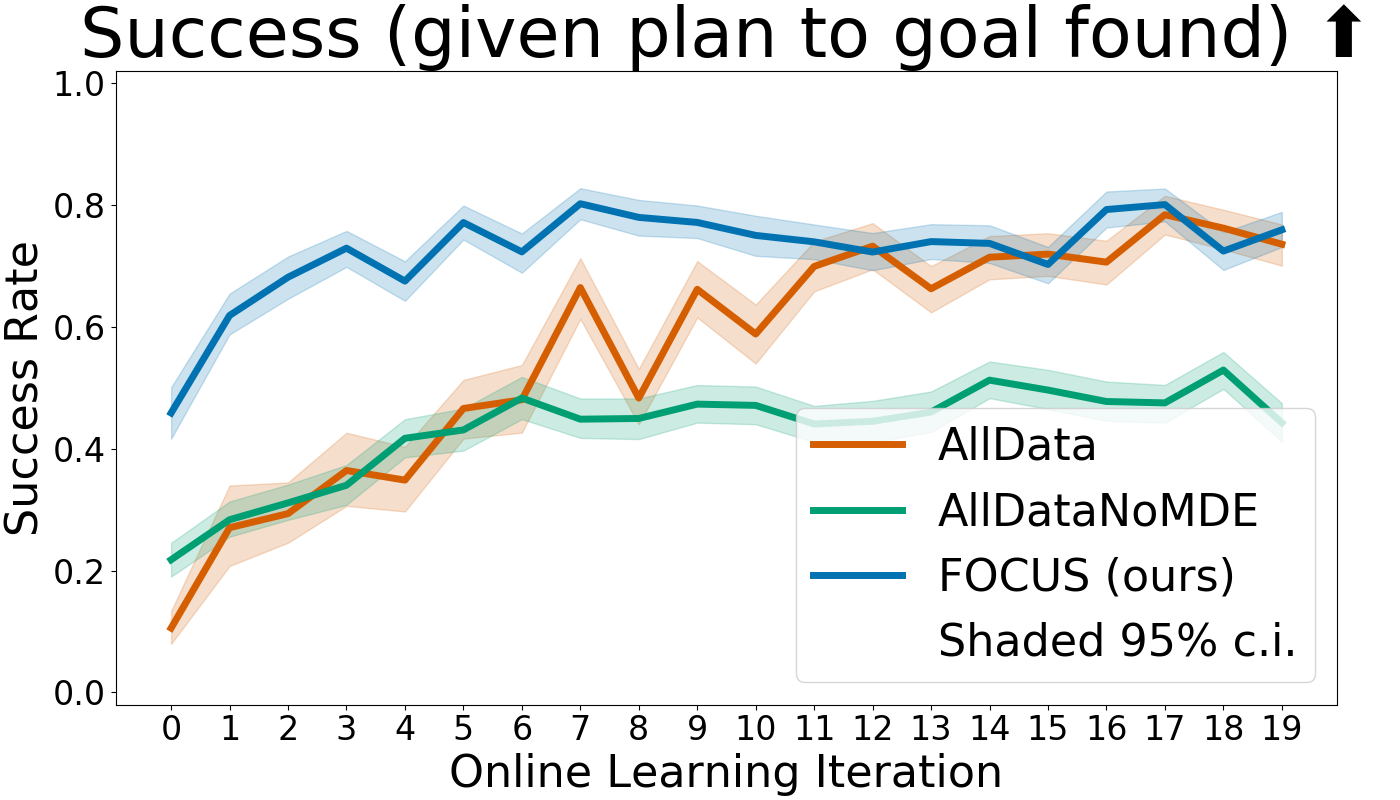} 
     \end{subfigure}
     \hfill
     \begin{subfigure}[b]{0.32\textwidth}
         \centering
         \includegraphics[width=\linewidth]{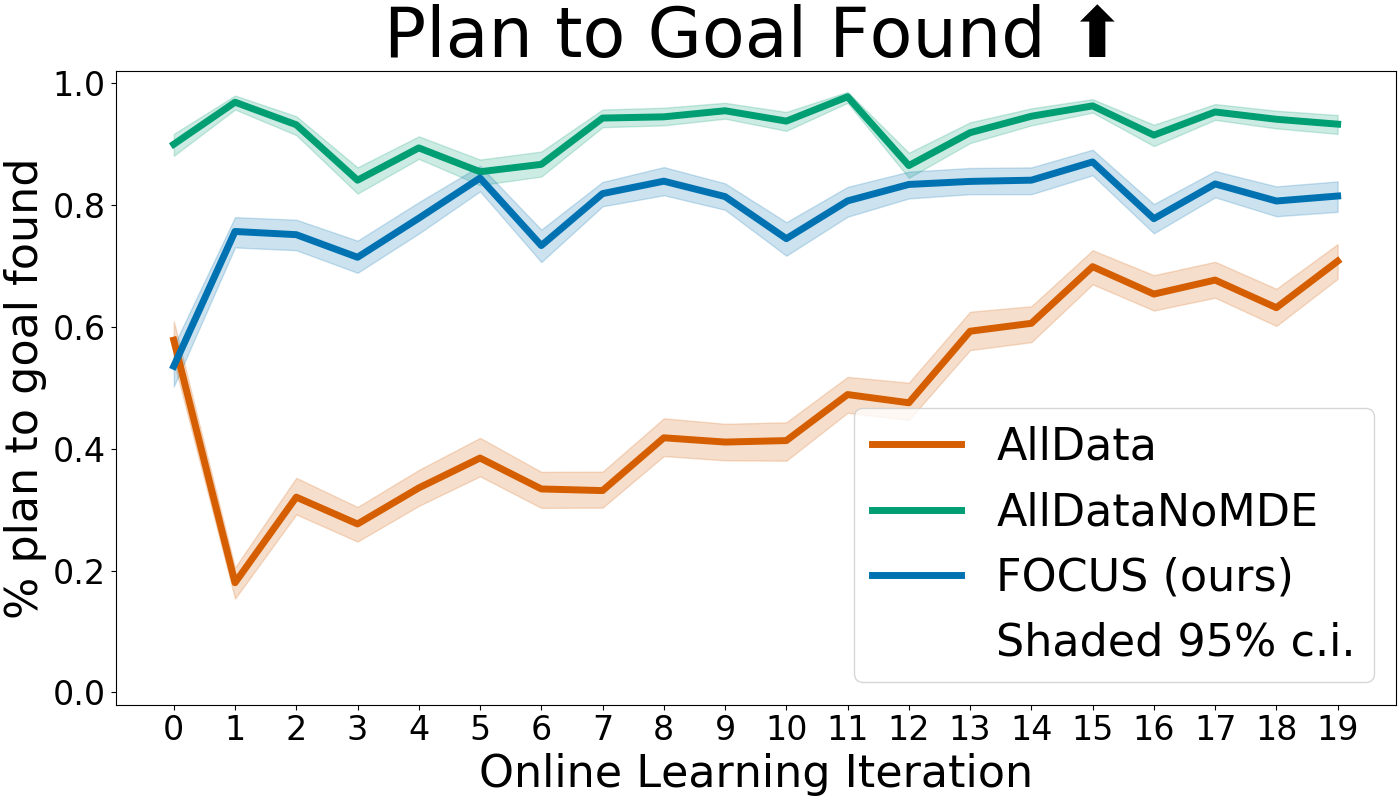}
     \end{subfigure}
     
    \caption{Post-learning evaluation of planning during test time in simulation: Metrics shown over the 20 iterations of online learning. If the planner does not find a plan that reaches the goal, the plan with the lowest distance to the goal is executed. We report (1) overall success, (2) success given that the plan reaches the goal, (3) and the percentage of plans that reach the goal. The shaded interval is the 95\% confidence interval of the mean, with the boot-strapping method used by Seaborn.}
    \label{fig:sim_results}
\end{figure*}

\begin{figure}
    \centering
    \includegraphics[width=\linewidth]{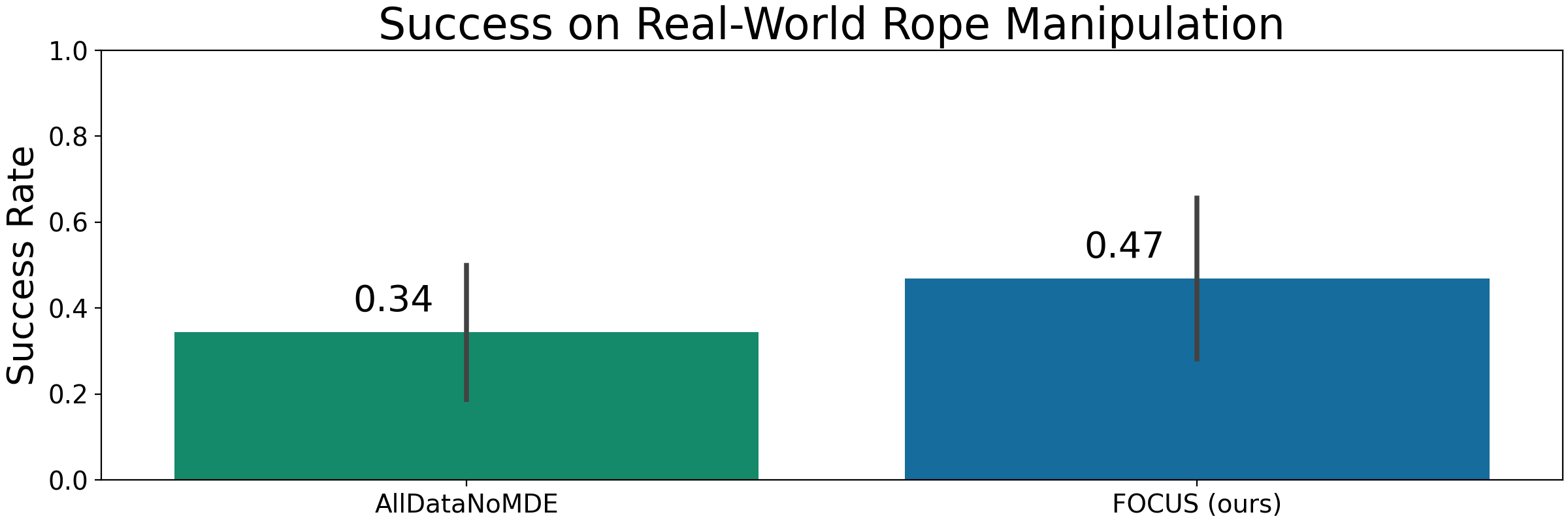}
    \caption{Success rate for the \textit{AllDataNoMDE} baseline (left) versus \FOCUS{} (right) for online adaptation to real-world bimanual rope manipulation. Error bars show the 95\% confidence interval.}
    \label{fig:real_rope_success}
\end{figure}

\textbf{Plant Watering: }
The goal in this task, illustrated in Figure~\ref{fig:waterscene} is to pour at least 75\% of the initial volume from a controlled container into a target container without spilling more than 2\%. The source environment is a variation from the SoftGym PourWater environment~\cite{lin2020softgym}. The controlled container can rotate about the z-axis. The state space is the 4-DOF $[x,y,z,\theta]$ pose of the controlled container, 3-DOF $[x,y,z]$ pose of the target container, control volume, and target volume. The action space is a target pose $[x_{des}, y_{des}, \theta_{des}]$ which is followed by a proportional controller. The distance function $\dist$ is a weighted sum of the distance in x, y, theta, and volume.

The target environment has triple the viscosity, a shorter container, and a plant in the target container. Although the agent can pour from above, that causes the water to splatter, which is more difficult to predict than the free-space pours of the target environment. Additionally, the box can collide with the plant, which is dissimilar to the source dynamics where there are no obstacles. The set $\DST$ contains free-space motions and pours, whereas collision with and pouring on the plant is not in $\DST$. We use $\softFilteringThrehsold=0.10$.

\subsection{Validating the Adaptation Method}
\label{sec:validating_adaptation}

We now evaluate whether the proposed adaptation method achieves lower prediction error on $\DST$. We start by creating validation sets that contain transitions not used for training, and which are known to be in $\DST$. For rope, this means transitions where the rope is in free space. For water, this means transitions that do not collide with or pour over the plant. We evaluate our method and two baselines, all starting from the same pre-trained model and adapting to the same dataset from the target environment. The baseline \textit{AllData} fine-tunes on all transitions with equal weights. The baseline \textit{LowInitialError} uses our weighting function, but computes the weights once using $\initialDynamics$ and does not re-compute them throughout training. Our method uses our proposed loss function (Equation \eqref{eq:adaptation_loss}) which re-computes the weights on each batch during training. We do not compare to domain randomization methods, which require knowledge of which physical parameters may vary and corresponding bounds.

For bimanual rope manipulation, the dataset contains 6288 transitions and the validation set contains 792 transitions. For plant watering, the training dataset contains 854 transitions and the validation set contains 130 transitions. The results are visualized in Figure~\ref{fig:validating}. In both experiments, the error of our method is statistically significantly lower than both baselines ($p<0.0001$).

Figure~\ref{fig:weights} demonstrates the intuition behind our adaptation method. In the center, we show histograms over time of the weights assigned to the transitions in the training dataset, for water and for rope. The distribution is initially unimodal since the weighting function when $j=0$ is soft, but as training progresses the distribution rapidly becomes bimodal, where most transitions are given a weight of 1, but some transitions are given a weight of 0. We show examples of these low and high weight transitions on either side. For rope manipulation, we found that the number of the transitions with prediction error below $\gamma$ increases from 52\% at epoch 1 to 80\% at epoch 20, which shows that the subset of data we train on grows. This explains why our method outperforms the \textit{LowInitialError} baseline, since that baseline is not making use of as much of the data as our method does. The presence of transitions with 0 weight (e.g. 20\% for the rope domain at epoch 20) shows that our method is not converging to training on all examples, which does not perform well based on the \textit{AllData} baseline.

\subsection{Online Learning Experiments}
\label{sec:online_learning_experiments}

We show that \FOCUS{} achieves higher task success with less data than baselines which fine-tune the dynamics on all available data. The first baseline, called \textit{AllDataNoMDE} does not use our proposed adaptation method and does not use MDEs when planning, making it a conventional online learning method. The second, called \textit{AllData} includes MDEs in planning but ablates our fine-tuning method. First, we evaluate our method in the rope manipulation domain on adaptation from one simulation to another (see Figure \ref{fig:sim_rope_envs}). We ran 20 iterations of online learning, where each iteration consists of 10 episodes for rope manipulation, and 27 episodes for plant watering ($\approx$6000 total transitions during learning). We then repeated this 10 times for each method/baseline with different random seeds.

After learning, we took the models saved after each learning iteration and ran 100 episodes of evaluation per method. To maximize the success rates of all methods, we use a longer timeout and do not allow random-accepts when planning. We also stop execution and replan if the error between the plan and the observed state exceeds a large threshold on model error (0.25).

The results of this experiment are summarized in Figure \ref{fig:sim_results}. In both tasks, the proposed method (\FOCUS{}) shows the highest data efficiency in achieving the goal. In the rope manipulation task, \textit{AllData}, never finds paths to the goal. This is because its dynamics are not sufficiently accurate, and so the MDE constraint makes the planning problem infeasible. Accurately learning the dynamics using \textit{AllData} or \textit{AllDataNoMDE} involves predicting the deformation of the rope on obstacles, which is challenging given a dataset of only a few thousand transitions. The boundary of $\DST$ is less extreme in the watering task, enabling both baselines to achieve high success rates, though using more data than \FOCUS{}. By inspecting the behavior of the methods, we found that our method quickly focuses on pours that do not interact with the plant, improving data efficiency.

\subsection{Real Robot Results}
\label{sec:real_rope_experiments}

To demonstrate how \FOCUS{} enables a robot to quickly learn a task in the real world, we performed a similar experiment to the first rope manipulation experiment, but on real robot hardware, where sensor and actuator noise are substantial factors (approximately \SI{5}{\centi\meter} of end-effector error). We use CDCPD2~\cite{CDCPD2} to track the rope state. The geometry of the car scene is approximated with primitive geometric shapes. We use the same source simulation environment as for the simulation rope experiment, but now the target environment is the real world. The robot should adapt the simulated free-space rope dynamics from simulation to the real world, despite the different real-world free-space dynamics and the fact that the rope can deform on the robots' arms or on the objects in the scene. Because perception error and actuation error are higher in the real world than in simulation, we use $\softFilteringThrehsold=0.2$.

We ran the online learning procedure with a single start configuration and a single goal region for one random seed and compare \FOCUS{} to the \textit{AllDataNoMDE} baseline, since \textit{AllDataNoMDE} performed best in simulation. After 15 iterations of learning, we freeze the models and evaluate task success 32 times. In our unoptimized implementation, planning and fine-tuning the models took comparable amounts of time, each taking between 1 and 30 minutes depending on task complexity/dataset size. A more advanced implementation would run fine-tuning in parallel with execution so that the robot is idle only while planning.

The success rates are shown in Figure \ref{fig:real_rope_success}. With \FOCUS{}, the robot successfully placed the rope under the engine 15/32 times, while \textit{AllDataNoMDE} succeeded 11/32 times. Failure modes for FOCUS include the rope getting pulled out of the robots' hands, getting too close and catching on obstacles, and failing to find plans that reach the goal. We find less improvement in the real world than in simulation, which may be due to perception and significant actuator error.

\section{CONCLUSION}

This paper studies the problem of adapting learned dynamics models to datasets that contain transitions where the dynamics are very different from the source environment. This type of domain mismatch is common in online dynamics learning settings, where the source dynamics are learned in simulation or on a simpler task. Traditional adaptation methods can fail in this setting because trying to fit data from regions of dissimilar dynamics leads to poor predictions even in regions where the source and target dynamics are similar.

Our key insight is to instead focus adaptation on regions where the source and target dynamics are similar. We propose an adaptation method that assigns high weight to transitions with low prediction error, and dynamically re-assigns weights during the course of training. The set of low-error transitions is initially a small set, but grows as training pulls down the prediction error for other similar transitions. We combine our adaptation method with prior work on planning with unreliable dynamics to make \FOCUS{}, a data-efficient online adaptation method. We demonstrate that \FOCUS{} can learn a bimanual rope manipulation task in simulation and in the real world, and achieves higher task success rates than baselines that attempt to fit all the training data.

\addtolength{\textheight}{0cm}   

\bibliographystyle{unsrt}
\bibliography{references}

\end{document}